\documentclass[11pt]{article}
\usepackage[final]{acl}
\AtBeginDocument{%
  }

\pdfoutput=1
\usepackage{times}
\usepackage{latexsym}
\usepackage[T1]{fontenc}
\usepackage{microtype}
\usepackage{graphicx}

\usepackage{amsmath}
\usepackage{float}
\usepackage{booktabs}
\usepackage{enumitem}

\usepackage{pdfrender}
\usepackage{algorithm}
\usepackage{algpseudocode}
\usepackage{pifont}
\usepackage{natbib}

\title{Improving Few-Shot Cross-Domain Named Entity Recognition by Instruction Tuning a Word-Embedding based Retrieval Augmented Large Language Model}

\author{Subhadip Nandi \\
  IIT Kanpur, India \\
  \texttt{subhadipn650@gmail.com} \\\And
  Neeraj Agrawal \\
  IISc Bangalore, India \\
  \texttt{aneeraj@iisc.ac.in} \\}
  
\begin{document}

\maketitle

\begin{abstract}
Few-Shot Cross-Domain NER is the process of leveraging knowledge from data-rich source domains to perform entity recognition on data-scarce target domains. Most previous state-of-the-art (SOTA) approaches use pre-trained language models (PLMs) for cross-domain NER. However, these models are often domain specific. To successfully use these models for new target domains, we need to modify either the model architecture or perform model fine-tuning using data from the new domains. Both of these result in the creation of entirely new NER models for each target domain which is infeasible for practical scenarios. Recently, several works have attempted to use LLMs to solve Few-Shot Cross-Domain NER. However, most of these are either too expensive for practical purposes or struggle to follow LLM prompt instructions. In this paper, we propose IF-WRANER (Instruction Finetuned Word-embedding based Retrieval Augmented large language model for Named Entity Recognition), a retrieval augmented LLM, finetuned for the NER task. By virtue of the regularization techniques used during LLM finetuning and the adoption of word-level embedding over sentence-level embedding during the retrieval of in-prompt examples, IF-WRANER is able to outperform previous SOTA Few-Shot Cross-Domain NER approaches. We have demonstrated the effectiveness of our model by benchmarking its performance on the open source CrossNER dataset, on which it shows more than 2\% F1 score improvement over the previous SOTA model. 
We have deployed the model for multiple customer care domains of an enterprise. Accurate entity prediction through IF-WRANER helps direct customers to automated workflows for the domains, thereby reducing escalations to human agents by almost 15\% and leading to millions of dollars in yearly savings for the company.

\end{abstract}

\pagenumbering{gobble}
\pagestyle{plain} 
\setcitestyle{authoryear,open={},close={}} 




\section{Introduction}

Named Entity Recognition (NER) (\citep{chinchor1997muc}) is a key process in information extraction, designed to identify and categorize entities in natural language into predefined entity types. Due to the large variations in entities and the way they are used across domains, NER has been a challenging task in NLP. Most traditional NER models require large volumes of labelled data for training (\citep{wang2022deepstruct}; \citep{yu2020named}; \citep{wang2020automated}; \citep{li2022unified}).
However, collecting large volumes of labeled data is both costly and time consuming. Therefore, we need a model that can perform NER on multiple domains with minimal labeled examples from that domain. To tackle this problem, several solutions have been proposed, which attempt to transfer knowledge from data rich source domain to perform NER on data-scarce target domain. 
This is referred to as Few-Shot Cross-Domain NER. 

The traditional way of solving this involves training PLMs with entity-tagged source domain data, followed by fine-tuning them on target domain data, thereby transferring knowledge from source to target domain. This approach fails to address the semantic gap that may exist between source and target domains. 
To address this, some previous studies utilize adding auxiliary objects (\citep{liu2020zero}; \citep{wang2020multi}) or designing new model architectures (\citep{jia2019cross}; \citep{liu2020zero}; \citep{jia2020multi}) to train with both source and target domain data. \citep{liu2112ner} leverages continued pretraining with target domain data for a better understanding of domain-specific data. Another line of research (\citep{zheng2022cross}; \citep{hu2022label}) focuses on modeling the label relationship across domains to improve label information transfer. Specifically, LANER (\citep{hu2022label}) utilizes an architecture to better leverage semantic relationships among domain labels to increase cross-domain performance.

Most Few-Shot Cross-Domain NER models however, have one or both of the following weaknesses:
\begin{itemize}[noitemsep,topsep=1pt,labelindent=0.3em,leftmargin=*]
\item Models like LANER (\citep{hu2022label}), have architectures that are very specific to the source and target domain pairs. Making these models work well for a new target domain requires tweaking the architecture. 
\item Other approaches require finetuning of model on target domain data. This is not feasible in real life scenarios due to computational resource and time crunch.
\end{itemize}
Our approach involves a single model architecture, finetuned only using entity tagged source domain data. For adaptation to target domain, our model does not require further fine-tuning.

Recently, with the advent of generative AI, many researchers have tried to use LLMs to solve the Few Shot Cross-Domain NER problem. GPT-NER (\citep{wang2023gpt}) and PromptNER (\citep{ashok2023promptner}) have experimented with different LLM prompting strategies for the task with varying degrees of success. GPT-NER further demonstrates that using the Retrieval Augmented Generation (RAG) (\citep{lewis2020retrieval}) framework to select the in-prompt examples further boosts NER performance. One common theme that has emerged from these works is that most of these approaches demonstrate good performance with GPT4 as the backbone LLM. Open source alternatives typically fall well short of SOTA performance as they do not seem to closely follow the prompt instructions. This is a serious problem for real world scenarios as the use of proprietary software like GPT4 can be cost-prohibitive, especially for applications operating at scale. In our approach, we finetune open source LLMs (\citep{touvron2023llama}), so that they can follow domain specific prompt instructions for the NER task. 

Like GPT-NER (\citep{wang2023gpt}), we too utilise the RAG framework for selecting in-prompt examples. We store labelled domain data and their vector embeddings in a vector store and extract relevant domain examples from the store during inference based on similarity between the inference query embedding and the embeddings stored. Most applications using the RAG framework use sentence level embeddings for similarity score calculations. In our work, we show that for the NER task, retrieving examples based on word-level embedding similarity performs much better than that based on sentence-level embedding similarity.

\section{Background and Related Work}
Traditional approaches to solve the NER problem typically fall into one of the two categories:
\begin{itemize}[noitemsep,topsep=1pt,labelindent=0.5em,leftmargin=*]
\item BERT-based (\citep{devlin2018bert}) models like BERT-Tagger, introduced by \citep{ding2021few} are built by adding a linear classifier on top of BERT such that each token in the sentence is classified into one of the pre-defined entity types. These models are generally trained with a cross-entropy objective.
\item Meta-learning approaches like ProtoBERT (\citep{snell2017prototypical}), NN-Shot (\citep{yang2020simple}) and Struct shot (\citep{yang2020simple}) derive a prototype for each entity type by computing the average of the contextual embeddings of the tokens that share the same entity type. Nearest neighbour algorithms are then used to classify each token of an input sentence into one of the entity types based on similarity between the token embedding and the embeddings of the prototypes.
\end{itemize}
While meta-learning based techniques fare better than BERT based models in the few-shot setting, they still require substantial amounts of data for creating representative prototypes. 

Cross-domain Named Entity Recognition (NER) algorithms (\citep{lin2018neural}; \citep{yang2018design}) help to address the issue of data scarcity in the target domain by leveraging data from source domains. One commonly used strategy to tackle Few-Shot Cross-Domain NER is multitask learning, which involves the use of auxiliary objects (\citep{liu2020zero}; \citep{wang2020multi}) or the development of fresh model structures (\citep{jia2019cross}; \citep{liu2020zero}; \citep{jia2020multi}). These methods aim to enhance the NER performance in the target domain by training on data from both the source and target domains. Another facet of cross-domain NER research concentrates on the transfer of label information across domains. As noted by \citep{zheng2022cross}, the relationship of labels can be represented as a probability distribution to facilitate the transfer of cross-domain knowledge in NER more effectively. \citep{hu2022label} suggests a method to capitalise on the semantic relationships between domains more efficiently by utilising previous labels (from the source) and the corresponding token. \citep{chen2023one} introduces collaborative prefix tuning as a solution to the cross-domain NER issue. Although prefix-tuning is considerably quicker than complete fine-tuning, it still requires the addition and modification of new model parameters. Unlike the above approaches, IF-WRANER does not necessitate architectural alterations or model fine-tuning/prefix-tuning to adapt to new target domains, making it more suitable for practical applications.



Large language models (LLMs) (\citep{mann2020language}; \citep{hoffmann2022training}) have demonstrated remarkable proficiency in in-context learning, where they can generate results for a new test input using only a handful of task-specific examples. Operating under the in-context learning framework, LLMs have yielded encouraging outcomes across a range of NLP tasks, including Machine Translation (MT) (\citep{vilar2022prompting,moslem2023adaptive}), Question Answering (QA) (\citep{robinson2022leveraging,li2022self}) etc.
For Few-Shot Cross Domain NER, PromptNER (\citep{ashok2023promptner}) and GPT-NER (\citep{wang2023gpt}) have utilized LLMs, achieving a performance level comparable to the industry benchmark through extensive prompt engineering. GPT-NER has redefined the NER problem from a sequence labeling task to a generation task that LLMs can easily adapt to. On the other hand, PromptNER uses the Chain-of-Thought Prompting technique, which offers a precise, adaptable, and user-friendly way to carry out Few-Shot NER and requires prompting the LLM only once. In our approach, we instruct the LLM to provide responses in a structured format and both the extraction of entities and their categorization into entity types happens in a single LLM call. 

Another recent innovation that has gained tremendous popularity and adoption post the advent of LLMs is Retrieval-Augmented Generation (RAG) (\citep{lewis2020retrieval}) with LLMs. For Few-Shot Cross-Domain NER, we show that using proprietary LLM like GPT4 with the RAG framework we are able to obtain results comparable to SOTA models. However, given the cost implications of using GPT4, we have developed a strategy to finetune open-source LLMs. With our approach of finetuning open-source LLM, coupled with regularization techniques and replacing the similarity between sentence level embeddings with similarity between word level embedding as criteria for selecting relevant examples using the retriever, we are able to achieve better performance on most domains compared to previous SOTA models.

\section{Methodology}
\subsection{Problem Definition}
Given a list of predefined entity types for a domain, and a sentence, the Named Entity Recognition (NER) task involves identifying sequences of words in the sentence as entities and categorizing them into correct entity types. With Few-Shot Cross Domain NER, we have the added restriction that the number of labeled examples for the domain in question (target domain) is small. However, we have sufficient labeled examples from another domain (source domain) which we can leverage for model building.

\subsection{Prompting LLM for NER} \label{prompt}
NER has historically been viewed as a sequence labeling task that assigns an entity type (partial assignment in-case of multi-word entities) to each word in a given sentence. With the advent of LLMs, many studies have tried to reformulate NER as a text generation task instead of a sequence labelling task. The format of the generated text can vary widely but can be broken into two categories at a high level. 
\begin{itemize}[noitemsep,topsep=1pt,labelindent=0.5em,leftmargin=*]
    \item The output is a dictionary with candidate entities (sequences of words in the sentence) as keys and their corresponding entity types as values.
    \item The output is a dictionary with all the entity types as keys and their corresponding entities (sequences of words in the sentence) as values.
\end{itemize}
We saw greater success with the second approach and decided to adopt it for our subsequent experiments. Most approaches that leverage LLMs to solve the NER task follow the general paradigm of in-context learning that includes prompt construction, followed by feeding it to the LLM, which then produces output in the format described in the prompt. The prompts for our experiments have the following format:
\begin{itemize}[noitemsep,topsep=1pt,labelindent=0.5em,leftmargin=*]
\item \textbf{Task Description:} Explains the task to the LLM which in our case looks like: ``You are a smart and intelligent Named Entity Recognition (NER) system. You will be provided with the definition of the entities to extract, the sentence from which to extract the entities and the format in which you are to display the output...''
\item \textbf{Entity Definitions:} Contains list of entity types for a particular domain and their respective definitions.
\item \textbf{Input Output Examples:} Top k examples of input and output pairs from corresponding domain data such as:\\
Input: Can i pick this up tomorrow\\
Output: \{product:[], time:[tomorrow] etc.\}
\item \textbf{User Query:} Sentence on which NER is to be performed.
\end{itemize}

\subsection{Retrieval Augmented Generation (RAG)}\label{rallm}
\begin{figure}[h]
  \centering
  \includegraphics[width=\linewidth]{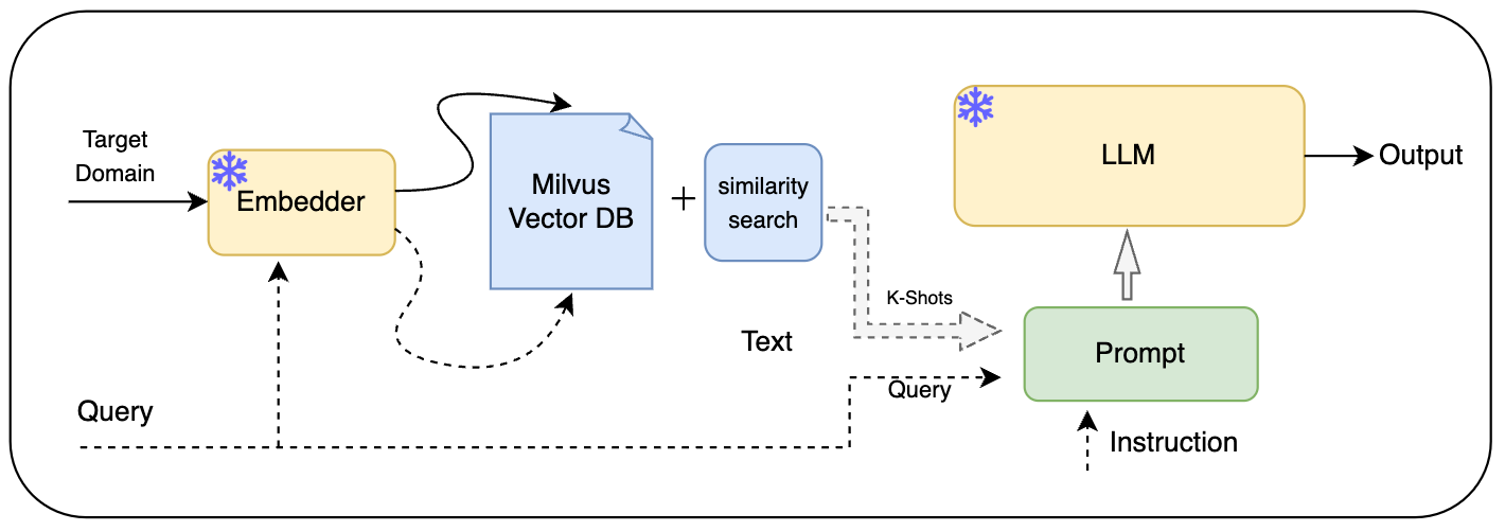}
  \captionsetup{font=small}
  \caption{\textmd{Retrieval Augmented Generation with LLM}}
  \label{llm_infer}
\end{figure}

With Retrieval Augmented Generation (RAG), instead of having the same hardcoded domain examples appended for every query, examples are selected dynamically based on their similarity with the input query. To achieve this, embeddings for domain examples are at first computed using a pre-trained universal embedder and the examples along with their embeddings are stored in a vector database. During inference, when a query comes in, its embedding is computed using the same embedder. Then similarity scores of the query embedding with all the embeddings stored in the vector db are computed and the top k most similar examples are selected and appended to the prompt, which is then sent to the LLM for response generation. 
The RAG framework is largely made up of two components
\begin{itemize}[noitemsep,topsep=1pt,labelindent=0.5em,leftmargin=*]
    \item \textbf{Retriever:} It is responsible for generating query embedding during inference using the embedder and also for extracting top k most similar examples from vector DB using a similarity function.
    \item \textbf{Generator:} The prompt, made up of the user query and the top k examples obtained by the retriever are passed along to the generator component (in our case an LLM) which is responsible for generating a response.
\end{itemize}
The complete RAG architecture is shown in Figure \ref{llm_infer}.
RAG with LLM yields good results for Few-Shot Cross-Domain NER when using proprietary LLMs like GPT4. However, most open source LLMs struggle to produce output in the format specified as part of prompt instruction. This becomes a challenge, because using GPT4 to perform NER for applications at scale can be extremely costly. Therefore, we need to finetune open source LLMs so that they can follow prompt instructions.

\subsection{Finetuning open-source LLMs for Few-Shot Cross-Domain NER}
\begin{figure}[h]
  \centering
  \includegraphics[width=\linewidth]{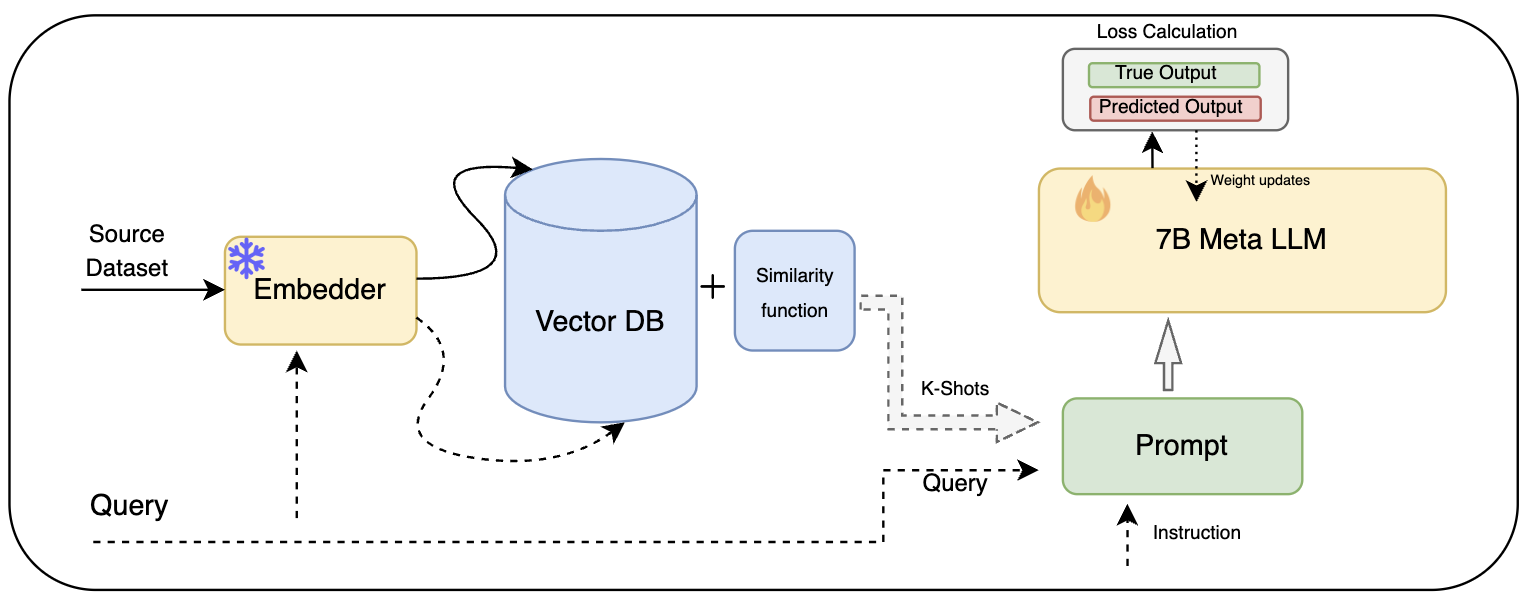}
  \captionsetup{font=small}
  \caption{\textmd{Finetune 7B Meta LLM on source domain data}}
  \label{finetune_fig}
  \vspace{-0.5cm}
\end{figure}

As per the Cross-Domain NER setting, we have a source domain which has enough entity tagged data. We finetune 7B Meta LLM on this source domain data. The purpose of this finetuning is not to teach the LLM about the source domain. Instead, finetuning accomplishes the task of teaching LLM to perform NER task and generate results in the format specified in the prompt.
The prompt content is the same as that in Section \ref{prompt}. We utilise the RAG framework during finetuning also, by storing a portion (around 500 examples) of the source domain data in a vector database and finetuning with the rest of the source domain examples. More details on this can be found in Section \ref{details}. Cross entropy loss, computed from the LLM output and ground truth, is used to finetune the LLM parameters. We employ LoRA (\citep{hu2021lora}) for this. The detailed finetuning process can be found in Figure \ref{finetune_fig}. 

Once finetuned on source domain data, we do not need to make any further changes to the model weights to adapt to different target domains. To evaluate model performance on any target domain, we simply store the labelled examples of that domain in a vector DB, and prompt the finetuned LLM in accordance with the RAG framework to generate outputs. 

\begin{figure*}[h]
  \centering
  \includegraphics[width=0.8\linewidth,scale=0.3]{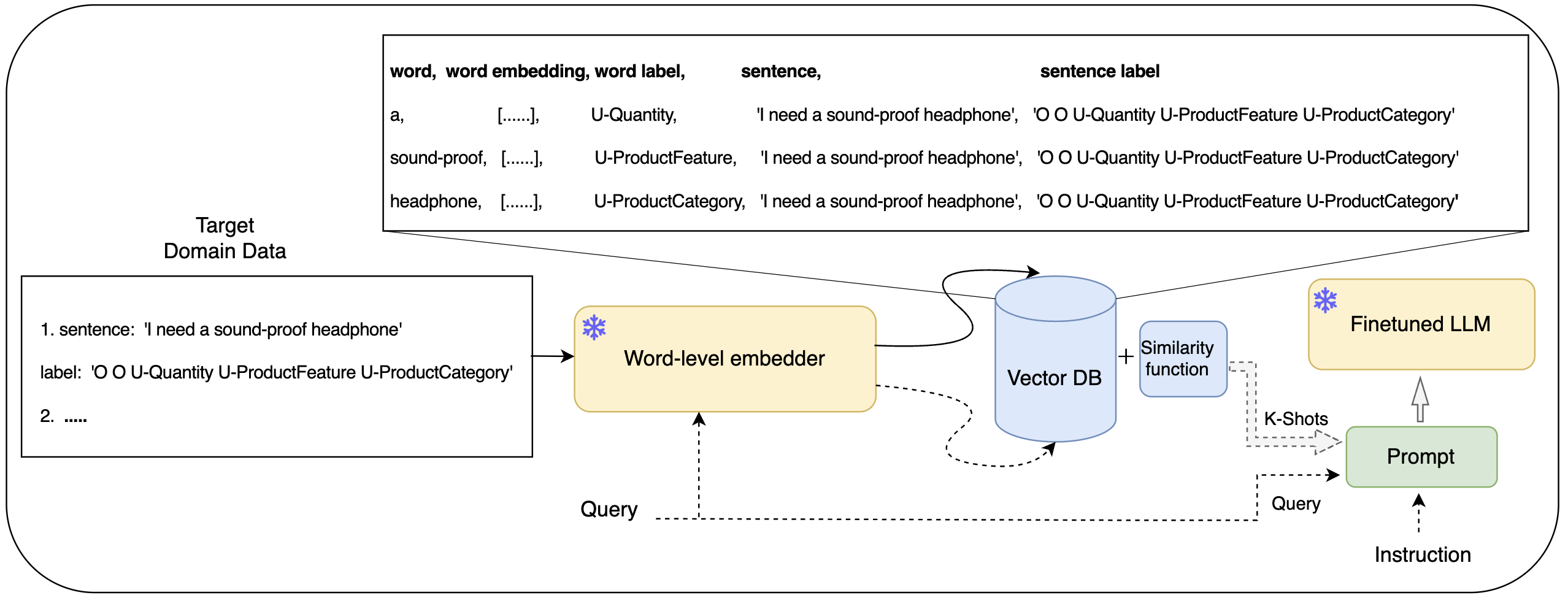}
  \captionsetup{font=small}
  \caption{\textmd{Using word-level embedding instead of sentence-level embedding}}
  \label{fig_word_level}
\end{figure*}
\subsection{Training Regularization}
While testing our model on target domain data, we found it to suffer from the problem of overfitting. As an example, for one of our target domains (Politics), our model tagged several politicians as ``Person'' instead of the entity type ``Politician'', despite the prompt instruction explicitly stating that only non-politicians were to be identified as ``Person''. This happens because the entity type ``Person'' was also part of the source domain and during finetuning, the model had memorized how to tag particular entity vaues as ``Person''. During evaluation, the LLM chose to ignore the instruction and continued to identify politicians as ``Person'' as learned during finetuning. In order to alleviate this problem, we introduce various kinds of noise during model finetuning. This prevents the model from memorizing entities for the different entity types and instead teaches the model to follow prompt instructions.
We applied the following regularization techniques:
\begin{itemize}[noitemsep,topsep=1pt,labelindent=0.5em,leftmargin=*]
    \item We duplicated a percentage of training examples and had some entity types randomly removed from both the input and output of those examples, which were then augmented to our training data. This ensures that the model is penalized when it predicts an entity type which is not part of the prompt, thus forcing it to learn to respect the prompt instruction.
    \item We randomly shuffled the order of entity types in the prompt for some examples. This prevents the model from memorizing the prompt and helps it achieve robustness against changes in the ordering of entity types in the prompt. 
\end{itemize}
A comparison of the relative contribution of the regularization techniques described has been shown in Section \ref{ablation_section} of Appendix.

\subsection{Using word-level embedding instead of sentence-level embedding}
NER is a word-level task that focuses more on local evidence rather than a sentence-level task, which is concerned with sentence-level semantics. Let us consider the following query sentence: ``I want to buy a 13-inch macbook from store''. We have two candidates sentences for adding as example in prompt:
``I want to buy a table from store'' and ``Show me a 15-inch macbook''. If we consider sentence-level embeddings to compute similarity, candidate 1 is closer to the query sentence. However, for the NER task, candidate 2 would be a much better example to have in the prompt as it contains a very similar entity to the example sentence. To resolve this, we retrieve examples based on word-level representations rather than sentence-level representations. Implementing this involves the following steps:
\begin{itemize}[noitemsep,topsep=1pt,labelindent=0.5em,leftmargin=*]
    \item Obtain contextualized word embedding for every entity tagged word across all sentences in domain data. This is done by passing the sentences through an encoder model (bge-base-en (\citep{bge_embedding})). Tokens corresponding to the same word are averaged to obtain embedding for every word in the sentence.
    \item We store each word embedding, the word itself, the corresponding sentence and sentence label in our vector DB.
    \item During inference, we obtain embedding for every word in input sentence in the same manner. 
    \item For every word, we find the top k closest matches from vector DB based on cosine similarity of the embeddings and extract the associated labelled examples. We end up with k X N examples where N is the number of words in the input sentence (after removing stop words), along with the word-level similarity scores.
    \item Among these, we select k unique examples with highest scores.
\end{itemize}

The process is shown in Figure \ref{fig_word_level}. The impact of using word embeddings instead of sentence embeddings for retrieval is shown in Appendix section \ref{word_not_sentence}.

\section{Experimental Setup}

\subsection{Dataset} \label{dataset}
We have developed our model for the customer care domain of an enterprise. However, given the proprietary nature of the data, we have not shared it here. To demonstrate the generalized nature of our  model, we have also conducted experiments using the open source CrossNER (\citep{liu2021crossner}) dataset. CrossNER contains separate datasets from five diverse domains, namely politics, natural science, music, AI, and literature. We adhere to the official splits for training, validation and test sets, the details of which can be found in Table \ref{dataset_table}. Models that report their results on the CrossNER dataset typically use a subset of the CoNLL 2003 (\citep{sang2003introduction}) dataset as source domain, also provided as part of the CrossNER dataset. We follow the same guidelines and use the subset of CoNLL 2003 as our source domain data.

\begin{table}[h] 
    \centering
    \scalebox{0.7}{
    \begin{tabular}
{p{3.2cm}|p{2cm}|p{1.8cm}|p{1.8cm}}
    \hline
    \textbf{Domain}&\textbf{No. of train examples}&\textbf{No. of dev examples}&\textbf{No. of test examples}\\  
    \hline
     Reuters&14987&3466&3684\\
     
     \hline
     Politics&200&541&651\\
     Natural Science&200&450&543\\
     Music&100&380&456\\
     Literature&100&400&416\\
     AI&100&350&431\\
    \hline   
    \end{tabular}}
    \captionsetup{font=small}
    \caption{\textmd{Dataset Statistics - Reuters from CoNLL 2003 is used as source domain. The rest of the domains from CrossNER dataset make up the target domains}}
    \label{dataset_table}
\end{table}

\subsection{Experiment Details} \label{details}
We use a 7B Meta LLM as the open source model for our work. We have also experimented with Mistral-7B (\citep{jiang2023mistral}), 13B Meta LLM and others, but based on the trade-off between model performance and model response latency, we found the 7B Meta LLM to be ideal for our use case. Detailed comparison of the models can be found in Appendix section \ref{open_source_llm}. 

As mentioned in Section \ref{dataset}, we use CoNNL 2003 dataset as our source domain and finetune our model using it. We randomly sample 500 examples from the training set, generate embeddings for these examples and store these embeddings along with the labelled examples in a vector DB. The rest of the training examples are used for finetuning the LLM. The validation set is used for the selection of LLM hyperparameters. While evaluating model performance on the target domains, the respective training and validation sets, along with their embeddings are stored in vector DB. Inference is performed on the test set and the data from vector DB serves as potential examples to be used in the prompt. 

Since full-finetuning of LLMs is resource intensive we use the Parameter Efficient Fine-Tuning (PEFT) technique LoRA (4-bit) (\citep{hu2021lora}) for finetuning our model.
AdamW (\citep{loshchilov2017decoupled}) optimizer with a learning rate of 2e-4 is used during the process.
For our RAG framework, we had to choose from a plethora of options for vector DBs, embeding models and similarity metrics. We ended up using bge-base-en (\citep{bge_embedding}),  Milvus DB (\citep{wang2021milvus}) and cosine similarity respectively, based on empirical results. IVF Flat indexing method is used for indexing data in the Milvus vector DB. We set the value of k (number of in-prompt examples) to 5 for our experiments based on validation dataset results. We used V100 GPU for finetuning our model on CoNLL 2003 data. The whole finetuning process takes around 50 minutes.

\section{Results}\label{results_section}
In table \ref{results_table} we have compared the performance of our model against previous SOTA Cross-Domain NER models on the 5 domains of the CrossNER dataset. Details of the previous SOTA models which we have taken as baselines for our work can be found in Section \ref{baselines} of Appendix. IF-WRANER outperforms most of the models by a significant margin. Only PromptNER with GPT 4 is close in performance to our model. PromptNER with GPT3.5 falls well short of SOTA performance. Among the non-LLM approaches, CP-NER performs the best. 
We use micro F1-score for performance comparison, the most common metric for evaluating NER models.(\citep{ma2016end}; \citep{lample2016neural}).

\begin{table*}[h]
    
    \centering
    \scalebox{0.85}{
    \begin{tabular}
{p{7.5cm}|p{1.3cm}|p{1.6cm}|p{1.2cm}|p{1.7cm}|p{1.1cm}|p{1.3cm}}
    \hline
    \textbf{Model}&\textbf{Politics}&\textbf{Natural Science}&\textbf{Music}&\textbf{Literature}&\textbf{AI}&\textbf{Average}\\  
    \hline
    BiLSTM-CRF (\citep{lample2016neural}) &56.60&49.97&44.79&43.03&43.56&47.59\\
    Coach (\citep{liu2020coach}) &61.50&52.09&51.66&48.35&45.15&51.75\\
    CROSS-DOMAIN LM (\citep{jia2019cross}) &68.44&64.31&63.56&59.59&53.70&61.92\\
    FLAIR (\citep{akbik2018contextual}) &69.54&64.71&65.60&61.35&52.48&62.73\\
    BARTNER (\citep{yan2021unified}) &69.90&65.14&65.35&58.93&53.00&62.46\\
    LIGHTNER (\citep{liu2020coach}) &72.78&66.74&72.28&65.17&35.82&62.56\\
    LST-NER (\citep{zheng2022cross}) &73.25&70.07&76.83&70.76&63.28&70.84\\
    LANER (\citep{hu2022label}) &74.06&71.83&78.78&71.11&65.79&72.31\\
    CP-NER (\citep{chen2023one}) &74.25&75.82&79.10&72.17&67.95&73.86\\
    \hline
    GPT-NER (\citep{wang2023gpt})                 &74.71&70.77&78.30&62.18&66.07&70.41\\
    PromptNER (GPT3.5) (\citep{ashok2023promptner}) &71.74&64.83&77.78&64.15&59.35&67.57\\
    PromptNER (GPT4) (\citep{ashok2023promptner}) &78.61&72.59&84.26&74.44&64.83&74.95\\
    \hline
    RAG + GPT4 using sentence embeddings &78.2&73.52&83.61&71.32&66.91&74.71\\
    RAG + GPT4 using word embeddings &78.63&73.95&84.25&74.68 & 68.19 & 75.94\\
    \textbf{IF-WRANER} (ours)        &\textbf{79.8}&\textbf{75.31}&\textbf{85.43}&\textbf{75.52}&\textbf{68.81}&\textbf{76.97}\\
    \hline         
    \end{tabular}}
\captionsetup{font=small}
\caption{\textmd{Comparisons of previous SOTA models for Cross-Domain NER and IF-WRANER in terms of F1 scores(\%) are provided. The Average indicates the average F1 score across five domains in the CrossNER benchmark}}
\label{results_table}
\end{table*}

\begin{table*}[t] 
  \scalebox{0.74}{
  \begin{tabular}{c | c  c  c |c | c | c}
    \toprule
    Model&\multicolumn{3}{c|}{Performance}&Latency (s)& QPS &Cost/month ($\$$)\\
    & CrossNER& Domain A& Domain B& &\\
    \midrule
    IF-WRANER (ours)     & 76.97& 83.72& 79.95& 2X& 1X & 1X\\
    Tiny-IF-WRANER (ours)  & 73.62& 79.64&76.46&  1X& 1X & 1X\\
    RAG with GPT 4 (our implementation)  & 74.71& 80.95&78.04&  4X& 1X & 120X\\
    PromptNER with GPT4(\citep{ashok2023promptner})   & 74.95& 81.15& 78.12& 4.2X& 1X & 120X\\
   \bottomrule
\end{tabular}}
\captionsetup{font=small}
\caption{\textmd{Comparison of model performance, latency, throughput and cost for IF-WRANER, Tiny-IF-WRANER, RAG with GPT4 and PromptNER. F1 score(\%) is used as model performance metric. Latency, throughput and cost are expressed in seconds(s), queries per second(QPS) and USD respectively. The cost for IF-WRANER and Tiny-IF-WRANER is the cost of using A100 GPUs while the cost of PromptNER is the cost of calling GPT4 openai endpoint}}
 \label{deployment}
\end{table*}

\section{Model Deployment}
\begin{table}[h]
  \scalebox{0.75}{
  \begin{tabular}{p{0.22\linewidth} | p{0.33\linewidth} | p{0.3\linewidth} | p{0.25\linewidth}}
    \toprule
    \textbf{Domain}&\textbf{Data size in vector DB}&\textbf{Test Data Size}&\textbf{Number of entity types}\\
    \midrule
    Domain A & 200 & 400 & 8\\
    Domain B & 230 & 500 & 15\\
    \bottomrule
  \end{tabular}}
 \caption{\textmd{Characteristics of proprietary datasets}}
 \label{enterprise_data}
\end{table}

With IF-WRANER, we have built a model that new domains can use off the shelf, simply by adding entity type definitions and a few labelled examples from the respective domains. We use the tensorrt framework on Triton Inference Server (\citep{tillet2019triton}) for serving our model. Depending on the traffic and latency requirements for each domain, we create separate instances of our model and serve them with Triton.

Using IF-WRANER we are able to achieve reasonable latency and  throughput numbers on A100 GPUs. 
Some domains however, have very low latency requirements and as per our experiments, a 7B-parameter IF-WRANER cannot meet these latency requirements. For such domains, we create a new model with Tinyllama (\citep{zhang2024tinyllama}) as the base LLM. Tinyllama is a 1.1B model with the same architecture and tokenizer as 7B Meta LLM, pretrained on 3 trillion tokens. We finetune Tinyllama in exactly the same way as before. This finetuned Tinyllama, Tiny-IF-WRANER, is able to serve the domains with very low latency requirement. As expected, due to its smaller model size, Tiny-IF-WRANER suffers from a drop in F1-score. In Table \ref{deployment}, we have compared our models (IF-WRANER and Tiny-IF-WRANER) with GPT4 based models in terms of performance, latency, throughput and cost. We see that our models are able to serve the same throughput at much lower latencies and cost. The domain with low latency requirement is represented as domain A. We use tiny-IF-WRANER to serve its users. Domain B, without such a requirement, uses IF-WRANER. Both domain A and domain B are customer care domains of an e-commerce enterprise. Due to the proprietary nature of the domains, we cannot make their datasets available. However, we have shared some characteristics of the datasets in Table \ref{enterprise_data}. 


\section{Conclusion}
In this work, we have introduced IF-WRANER, a retrieval augmented instruction following LLM, that outperforms SOTA models for Few-Shot Cross-Domain NER. Unlike many of the models developed for Cross-Domain NER, we do not need to finetune or make structural modifications to our model to adapt to new domains. Also, IF-WRANER manages to attain SOTA performance using non-proprietary LLM, making it much more cost effective compared to proprietary LLM based Cross-Domain NER models. Our model is flexible and can be easily used by end users with no technical expertise. All they have to do is provide definitions for their domain's entity types and a few labelled examples. For serving domains with very low latency requirements, we have proposed tiny-IF-WRANER which uses Tinyllama instead of 7B Meta LLM as its base LLM.


\bibliography{references}

\begin{thebibliography}{49}
\expandafter\ifx\csname natexlab\endcsname\relax\def\natexlab#1{#1}\fi

\bibitem[{MPT()}]{MPT7b}

\newblock Introducing mpt-7b: A new standard for open-source, commercially usable llms.
\newblock \url{https://www.databricks.com/blog/mpt-7b}.

\bibitem[{Akbik et~al.(2018)Akbik, Blythe, and Vollgraf}]{akbik2018contextual}
Alan Akbik, Duncan Blythe, and Roland Vollgraf. 2018.
\newblock Contextual string embeddings for sequence labeling.
\newblock In \emph{Proceedings of the 27th international conference on computational linguistics}, pages 1638--1649.

\bibitem[{Ashok and Lipton(2023)}]{ashok2023promptner}
Dhananjay Ashok and Zachary~C Lipton. 2023.
\newblock Promptner: Prompting for named entity recognition.
\newblock \emph{arXiv preprint arXiv:2305.15444}.

\bibitem[{Chen et~al.(2021)Chen, Li, Deng, Tan, Xu, Huang, Si, Chen, and Zhang}]{chen2021lightner}
Xiang Chen, Lei Li, Shumin Deng, Chuanqi Tan, Changliang Xu, Fei Huang, Luo Si, Huajun Chen, and Ningyu Zhang. 2021.
\newblock Lightner: A lightweight tuning paradigm for low-resource ner via pluggable prompting.
\newblock \emph{arXiv preprint arXiv:2109.00720}.

\bibitem[{Chen et~al.(2023)Chen, Li, Qiao, Zhang, Tan, Jiang, Huang, and Chen}]{chen2023one}
Xiang Chen, Lei Li, Shuofei Qiao, Ningyu Zhang, Chuanqi Tan, Yong Jiang, Fei Huang, and Huajun Chen. 2023.
\newblock One model for all domains: collaborative domain-prefix tuning for cross-domain ner.
\newblock \emph{arXiv preprint arXiv:2301.10410}.

\bibitem[{Chinchor and Robinson(1997)}]{chinchor1997muc}
Nancy Chinchor and Patricia Robinson. 1997.
\newblock Muc-7 named entity task definition.
\newblock In \emph{Proceedings of the 7th Conference on Message Understanding}, volume~29, pages 1--21.

\bibitem[{Devlin et~al.(2018)Devlin, Chang, Lee, and Toutanova}]{devlin2018bert}
Jacob Devlin, Ming-Wei Chang, Kenton Lee, and Kristina Toutanova. 2018.
\newblock Bert: Pre-training of deep bidirectional transformers for language understanding.
\newblock \emph{arXiv preprint arXiv:1810.04805}.

\bibitem[{Ding et~al.(2021)Ding, Xu, Chen, Wang, Han, Xie, Zheng, and Liu}]{ding2021few}
Ning Ding, Guangwei Xu, Yulin Chen, Xiaobin Wang, Xu~Han, Pengjun Xie, Hai-Tao Zheng, and Zhiyuan Liu. 2021.
\newblock Few-nerd: A few-shot named entity recognition dataset.
\newblock \emph{arXiv preprint arXiv:2105.07464}.

\bibitem[{Douze et~al.(2024)Douze, Guzhva, Deng, Johnson, Szilvasy, Mazar{\'e}, Lomeli, Hosseini, and J{\'e}gou}]{douze2024faiss}
Matthijs Douze, Alexandr Guzhva, Chengqi Deng, Jeff Johnson, Gergely Szilvasy, Pierre-Emmanuel Mazar{\'e}, Maria Lomeli, Lucas Hosseini, and Herv{\'e} J{\'e}gou. 2024.
\newblock The faiss library.
\newblock \emph{arXiv preprint arXiv:2401.08281}.

\bibitem[{Hoffmann et~al.(2022)Hoffmann, Borgeaud, Mensch, Buchatskaya, Cai, Rutherford, Casas, Hendricks, Welbl, Clark et~al.}]{hoffmann2022training}
Jordan Hoffmann, Sebastian Borgeaud, Arthur Mensch, Elena Buchatskaya, Trevor Cai, Eliza Rutherford, Diego de~Las Casas, Lisa~Anne Hendricks, Johannes Welbl, Aidan Clark, et~al. 2022.
\newblock Training compute-optimal large language models.
\newblock \emph{arXiv preprint arXiv:2203.15556}.

\bibitem[{Hu et~al.(2021)Hu, Shen, Wallis, Allen-Zhu, Li, Wang, Wang, and Chen}]{hu2021lora}
Edward~J Hu, Yelong Shen, Phillip Wallis, Zeyuan Allen-Zhu, Yuanzhi Li, Shean Wang, Lu~Wang, and Weizhu Chen. 2021.
\newblock Lora: Low-rank adaptation of large language models.
\newblock \emph{arXiv preprint arXiv:2106.09685}.

\bibitem[{Hu et~al.(2022)Hu, Zhao, Guo, Wan, and Chang}]{hu2022label}
Jinpeng Hu, He~Zhao, Dan Guo, Xiang Wan, and Tsung-Hui Chang. 2022.
\newblock A label-aware autoregressive framework for cross-domain ner.
\newblock In \emph{Findings of the Association for Computational Linguistics: NAACL 2022}, pages 2222--2232.

\bibitem[{Jia et~al.(2019)Jia, Liang, and Zhang}]{jia2019cross}
Chen Jia, Xiaobo Liang, and Yue Zhang. 2019.
\newblock Cross-domain ner using cross-domain language modeling.
\newblock In \emph{Proceedings of the 57th annual meeting of the association for computational linguistics}, pages 2464--2474.

\bibitem[{Jia and Zhang(2020)}]{jia2020multi}
Chen Jia and Yue Zhang. 2020.
\newblock Multi-cell compositional lstm for ner domain adaptation.
\newblock In \emph{Proceedings of the 58th annual meeting of the association for computational linguistics}, pages 5906--5917.

\bibitem[{Jiang et~al.(2023)Jiang, Sablayrolles, Mensch, Bamford, Chaplot, Casas, Bressand, Lengyel, Lample, Saulnier et~al.}]{jiang2023mistral}
Albert~Q Jiang, Alexandre Sablayrolles, Arthur Mensch, Chris Bamford, Devendra~Singh Chaplot, Diego de~las Casas, Florian Bressand, Gianna Lengyel, Guillaume Lample, Lucile Saulnier, et~al. 2023.
\newblock Mistral 7b.
\newblock \emph{arXiv preprint arXiv:2310.06825}.

\bibitem[{Lample et~al.(2016)Lample, Ballesteros, Subramanian, Kawakami, and Dyer}]{lample2016neural}
Guillaume Lample, Miguel Ballesteros, Sandeep Subramanian, Kazuya Kawakami, and Chris Dyer. 2016.
\newblock Neural architectures for named entity recognition.
\newblock \emph{arXiv preprint arXiv:1603.01360}.

\bibitem[{Le~Scao et~al.(2023)Le~Scao, Fan, Akiki, Pavlick, Ili{\'c}, Hesslow, Castagn{\'e}, Luccioni, Yvon, Gall{\'e} et~al.}]{le2023bloom}
Teven Le~Scao, Angela Fan, Christopher Akiki, Ellie Pavlick, Suzana Ili{\'c}, Daniel Hesslow, Roman Castagn{\'e}, Alexandra~Sasha Luccioni, Fran{\c{c}}ois Yvon, Matthias Gall{\'e}, et~al. 2023.
\newblock Bloom: A 176b-parameter open-access multilingual language model.

\bibitem[{Lewis et~al.(2020)Lewis, Perez, Piktus, Petroni, Karpukhin, Goyal, K{\"u}ttler, Lewis, Yih, Rockt{\"a}schel et~al.}]{lewis2020retrieval}
Patrick Lewis, Ethan Perez, Aleksandra Piktus, Fabio Petroni, Vladimir Karpukhin, Naman Goyal, Heinrich K{\"u}ttler, Mike Lewis, Wen-tau Yih, Tim Rockt{\"a}schel, et~al. 2020.
\newblock Retrieval-augmented generation for knowledge-intensive nlp tasks.
\newblock \emph{Advances in Neural Information Processing Systems}, 33:9459--9474.

\bibitem[{Li et~al.(2022{\natexlab{a}})Li, Fei, Liu, Wu, Zhang, Teng, Ji, and Li}]{li2022unified}
Jingye Li, Hao Fei, Jiang Liu, Shengqiong Wu, Meishan Zhang, Chong Teng, Donghong Ji, and Fei Li. 2022{\natexlab{a}}.
\newblock Unified named entity recognition as word-word relation classification.
\newblock In \emph{Proceedings of the AAAI Conference on Artificial Intelligence}, volume~36, pages 10965--10973.

\bibitem[{Li et~al.(2022{\natexlab{b}})Li, Zhang, and Zhao}]{li2022self}
Junlong Li, Zhuosheng Zhang, and Hai Zhao. 2022{\natexlab{b}}.
\newblock Self-prompting large language models for open-domain qa.
\newblock \emph{arXiv preprint arXiv:2212.08635}.

\bibitem[{Lin and Lu(2018)}]{lin2018neural}
Bill~Yuchen Lin and Wei Lu. 2018.
\newblock Neural adaptation layers for cross-domain named entity recognition.
\newblock \emph{arXiv preprint arXiv:1810.06368}.

\bibitem[{Liu et~al.()Liu, Jiang, Hu, Shi, and Fung}]{liu2112ner}
Z~Liu, F~Jiang, Y~Hu, C~Shi, and P~Fung.
\newblock Ner-bert: A pre-trained model for low-resource entity tagging. arxiv 2021.
\newblock \emph{arXiv preprint arXiv:2112.00405}.

\bibitem[{Liu et~al.(2020{\natexlab{a}})Liu, Winata, and Fung}]{liu2020zero}
Zihan Liu, Genta~Indra Winata, and Pascale Fung. 2020{\natexlab{a}}.
\newblock Zero-resource cross-domain named entity recognition.
\newblock \emph{arXiv preprint arXiv:2002.05923}.

\bibitem[{Liu et~al.(2020{\natexlab{b}})Liu, Winata, Xu, and Fung}]{liu2020coach}
Zihan Liu, Genta~Indra Winata, Peng Xu, and Pascale Fung. 2020{\natexlab{b}}.
\newblock Coach: A coarse-to-fine approach for cross-domain slot filling.
\newblock \emph{arXiv preprint arXiv:2004.11727}.

\bibitem[{Liu et~al.(2021)Liu, Xu, Yu, Dai, Ji, Cahyawijaya, Madotto, and Fung}]{liu2021crossner}
Zihan Liu, Yan Xu, Tiezheng Yu, Wenliang Dai, Ziwei Ji, Samuel Cahyawijaya, Andrea Madotto, and Pascale Fung. 2021.
\newblock Crossner: Evaluating cross-domain named entity recognition.
\newblock In \emph{Proceedings of the AAAI Conference on Artificial Intelligence}, volume~35, pages 13452--13460.

\bibitem[{Loshchilov and Hutter(2017)}]{loshchilov2017decoupled}
Ilya Loshchilov and Frank Hutter. 2017.
\newblock Decoupled weight decay regularization.
\newblock \emph{arXiv preprint arXiv:1711.05101}.

\bibitem[{Ma and Hovy(2016)}]{ma2016end}
Xuezhe Ma and Eduard Hovy. 2016.
\newblock End-to-end sequence labeling via bi-directional lstm-cnns-crf.
\newblock \emph{arXiv preprint arXiv:1603.01354}.

\bibitem[{Mann et~al.(2020)Mann, Ryder, Subbiah, Kaplan, Dhariwal, Neelakantan, Shyam, Sastry, Askell, Agarwal et~al.}]{mann2020language}
Ben Mann, N~Ryder, M~Subbiah, J~Kaplan, P~Dhariwal, A~Neelakantan, P~Shyam, G~Sastry, A~Askell, S~Agarwal, et~al. 2020.
\newblock Language models are few-shot learners.
\newblock \emph{arXiv preprint arXiv:2005.14165}.

\bibitem[{Moslem et~al.(2023)Moslem, Haque, Kelleher, and Way}]{moslem2023adaptive}
Yasmin Moslem, Rejwanul Haque, John~D Kelleher, and Andy Way. 2023.
\newblock Adaptive machine translation with large language models.
\newblock \emph{arXiv preprint arXiv:2301.13294}.

\bibitem[{Robinson et~al.(2022)Robinson, Rytting, and Wingate}]{robinson2022leveraging}
Joshua Robinson, Christopher~Michael Rytting, and David Wingate. 2022.
\newblock Leveraging large language models for multiple choice question answering.
\newblock \emph{arXiv preprint arXiv:2210.12353}.

\bibitem[{Sainz et~al.(2023)Sainz, Garc{\'\i}a-Ferrero, Agerri, de~Lacalle, Rigau, and Agirre}]{sainz2023gollie}
Oscar Sainz, Iker Garc{\'\i}a-Ferrero, Rodrigo Agerri, Oier~Lopez de~Lacalle, German Rigau, and Eneko Agirre. 2023.
\newblock Gollie: Annotation guidelines improve zero-shot information-extraction.
\newblock \emph{arXiv preprint arXiv:2310.03668}.

\bibitem[{Sang and De~Meulder(2003)}]{sang2003introduction}
Erik~F Sang and Fien De~Meulder. 2003.
\newblock Introduction to the conll-2003 shared task: Language-independent named entity recognition.
\newblock \emph{arXiv preprint cs/0306050}.

\bibitem[{Snell et~al.(2017)Snell, Swersky, and Zemel}]{snell2017prototypical}
Jake Snell, Kevin Swersky, and Richard Zemel. 2017.
\newblock Prototypical networks for few-shot learning.
\newblock \emph{Advances in neural information processing systems}, 30.

\bibitem[{Tillet et~al.(2019)Tillet, Kung, and Cox}]{tillet2019triton}
Philippe Tillet, Hsiang-Tsung Kung, and David Cox. 2019.
\newblock Triton: an intermediate language and compiler for tiled neural network computations.
\newblock In \emph{Proceedings of the 3rd ACM SIGPLAN International Workshop on Machine Learning and Programming Languages}, pages 10--19.

\bibitem[{Touvron et~al.(2023)Touvron, Martin, Stone, Albert, Almahairi, Babaei, Bashlykov, Batra, Bhargava, Bhosale et~al.}]{touvron2023llama}
Hugo Touvron, Louis Martin, Kevin Stone, Peter Albert, Amjad Almahairi, Yasmine Babaei, Nikolay Bashlykov, Soumya Batra, Prajjwal Bhargava, Shruti Bhosale, et~al. 2023.
\newblock Llama 2: Open foundation and fine-tuned chat models.
\newblock \emph{arXiv preprint arXiv:2307.09288}.

\bibitem[{Vilar et~al.(2022)Vilar, Freitag, Cherry, Luo, Ratnakar, and Foster}]{vilar2022prompting}
David Vilar, Markus Freitag, Colin Cherry, Jiaming Luo, Viresh Ratnakar, and George Foster. 2022.
\newblock Prompting palm for translation: Assessing strategies and performance.
\newblock \emph{arXiv preprint arXiv:2211.09102}.

\bibitem[{Wang et~al.(2022)Wang, Liu, Chen, Hong, Tang, and Song}]{wang2022deepstruct}
Chenguang Wang, Xiao Liu, Zui Chen, Haoyun Hong, Jie Tang, and Dawn Song. 2022.
\newblock Deepstruct: Pretraining of language models for structure prediction.
\newblock \emph{arXiv preprint arXiv:2205.10475}.

\bibitem[{Wang et~al.(2021)Wang, Yi, Guo, Jin, Xu, Li, Wang, Guo, Li, Xu et~al.}]{wang2021milvus}
Jianguo Wang, Xiaomeng Yi, Rentong Guo, Hai Jin, Peng Xu, Shengjun Li, Xiangyu Wang, Xiangzhou Guo, Chengming Li, Xiaohai Xu, et~al. 2021.
\newblock Milvus: A purpose-built vector data management system.
\newblock In \emph{Proceedings of the 2021 International Conference on Management of Data}, pages 2614--2627.

\bibitem[{Wang et~al.(2020{\natexlab{a}})Wang, Kulkarni, and Preo{\c{t}}iuc-Pietro}]{wang2020multi}
Jing Wang, Mayank Kulkarni, and Daniel Preo{\c{t}}iuc-Pietro. 2020{\natexlab{a}}.
\newblock Multi-domain named entity recognition with genre-aware and agnostic inference.
\newblock In \emph{Proceedings of the 58th annual meeting of the association for computational linguistics}, pages 8476--8488.

\bibitem[{Wang et~al.(2023)Wang, Sun, Li, Ouyang, Wu, Zhang, Li, and Wang}]{wang2023gpt}
Shuhe Wang, Xiaofei Sun, Xiaoya Li, Rongbin Ouyang, Fei Wu, Tianwei Zhang, Jiwei Li, and Guoyin Wang. 2023.
\newblock Gpt-ner: Named entity recognition via large language models.
\newblock \emph{arXiv preprint arXiv:2304.10428}.

\bibitem[{Wang et~al.(2020{\natexlab{b}})Wang, Jiang, Bach, Wang, Huang, Huang, and Tu}]{wang2020automated}
Xinyu Wang, Yong Jiang, Nguyen Bach, Tao Wang, Zhongqiang Huang, Fei Huang, and Kewei Tu. 2020{\natexlab{b}}.
\newblock Automated concatenation of embeddings for structured prediction.
\newblock \emph{arXiv preprint arXiv:2010.05006}.

\bibitem[{Xiao et~al.(2023)Xiao, Liu, Zhang, and Muennighoff}]{bge_embedding}
Shitao Xiao, Zheng Liu, Peitian Zhang, and Niklas Muennighoff. 2023.
\newblock \href {http://arxiv.org/abs/2309.07597} {C-pack: Packaged resources to advance general chinese embedding}.

\bibitem[{Yan et~al.(2021)Yan, Gui, Dai, Guo, Zhang, and Qiu}]{yan2021unified}
Hang Yan, Tao Gui, Junqi Dai, Qipeng Guo, Zheng Zhang, and Xipeng Qiu. 2021.
\newblock A unified generative framework for various ner subtasks.
\newblock \emph{arXiv preprint arXiv:2106.01223}.

\bibitem[{Yang et~al.(2018)Yang, Liang, and Zhang}]{yang2018design}
Jie Yang, Shuailong Liang, and Yue Zhang. 2018.
\newblock Design challenges and misconceptions in neural sequence labeling.
\newblock \emph{arXiv preprint arXiv:1806.04470}.

\bibitem[{Yang and Katiyar(2020)}]{yang2020simple}
Yi~Yang and Arzoo Katiyar. 2020.
\newblock Simple and effective few-shot named entity recognition with structured nearest neighbor learning.
\newblock \emph{arXiv preprint arXiv:2010.02405}.

\bibitem[{Yu et~al.(2020)Yu, Bohnet, and Poesio}]{yu2020named}
Juntao Yu, Bernd Bohnet, and Massimo Poesio. 2020.
\newblock Named entity recognition as dependency parsing.
\newblock \emph{arXiv preprint arXiv:2005.07150}.

\bibitem[{Zaratiana et~al.(2023)Zaratiana, Tomeh, Holat, and Charnois}]{zaratiana2023gliner}
Urchade Zaratiana, Nadi Tomeh, Pierre Holat, and Thierry Charnois. 2023.
\newblock Gliner: Generalist model for named entity recognition using bidirectional transformer.
\newblock \emph{arXiv preprint arXiv:2311.08526}.

\bibitem[{Zhang et~al.(2024)Zhang, Zeng, Wang, and Lu}]{zhang2024tinyllama}
Peiyuan Zhang, Guangtao Zeng, Tianduo Wang, and Wei Lu. 2024.
\newblock Tinyllama: An open-source small language model.
\newblock \emph{arXiv preprint arXiv:2401.02385}.

\bibitem[{Zheng et~al.(2022)Zheng, Chen, and Ma}]{zheng2022cross}
Junhao Zheng, Haibin Chen, and Qianli Ma. 2022.
\newblock Cross-domain named entity recognition via graph matching.
\newblock In \emph{Findings of the Association for Computational Linguistics: ACL 2022}, pages 2670--2680.

\end{thebibliography}

\section{Appendix}

\subsection{Baselines}\label{baselines}  
To evaluate the effectiveness of the proposed
method, we compare it with several baselines, including:
\begin{itemize}[nosep,noitemsep,topsep=1pt,labelindent=0.5em,leftmargin=*]
    \item COACH (\citep{liu2020coach}): Utilizes patterns of slot entities and combines the features for each slot entity in order to improve entity type predictions.
    \item CROSS-DOMAIN LM (\citep{jia2019cross}): Utilizes a parameter generation network to merge crossdomain language modeling with NER.
    \item FLAIR (\citep{akbik2018contextual}): Utilizes the internal states of a character-level language model to generate contextual string embeddings, which are integrated into the NER model.
    \item BARTNER (\citep{yan2021unified}):Uses the pre-trained BART model to generate entity spans, treating the NER task as a sequence generation problem.
    \item LST-NER (\citep{zheng2022cross}): Models the relationship between labels as a probability distribution and builds label graphs in both the source and target label spaces for cross-domain NER tasks.
    \item LANER (\citep{hu2022label}): Uses a new approach for cross-domain named entity recognition by utilizing an autoregressive framework to strengthen the connection between labels and tokens.
    \item LIGHTNER (\citep{chen2021lightner}): Utilizes a pluggable prompting method to improve NER performance in low-resource settings.
    \item CP-NER (\citep{chen2023one}): Utilizes collaborative prefix tuning to learn domain-specific prefixes for flexible NER execution
    \item GPT-NER (\citep{wang2023gpt}): Redefines NER from a sequence labeling task to a generation task that LLMs can perform easily.
    \item PromptNER (\citep{ashok2023promptner}): Uses the Chain-of-Thought Prompting to perform NER.
\end{itemize}
Recently GoLLIE (\citep{sainz2023gollie}) and GLINER (\citep{zaratiana2023gliner}) have demonstrated SOTA performance on CrossNER dataset for the zero shot setting. However, they have not neither evaluated their models on the few-shot setting for CrossNER nor provided an easy way to do so. Therefore, we have not included them as baselines for our work.

\subsection{Ablation Study}\label{ablation_section}
We have performed an ablation study to compare the contributions of the different regularization techniques towards our model's performance. As shown in Table \ref{ablation}, regularization by augmenting examples with randomly removed entity types makes the most significant contribution to our model's performance.
\begin{table}[h]
  \scalebox{0.75}{
  \begin{tabular}{p{0.9\linewidth} | p{0.3\linewidth}}
    \toprule
    \textbf{Model}&\textbf{F1 score (\%)}\\
    \midrule
    base finetuned model  & 73.65\\
    base finetuned model + entity types removed  & 76.34\\
    base finetuned model + entity types shuffled  & 74.41\\
    base finetuned model + both & 76.97\\
   \bottomrule
  \end{tabular}}
 \caption{\textmd{Contribution of different regularization techniques towards model performance improvement}}
 \label{ablation}
\end{table}
\subsection{Effect of replacing sentence-level embedding with word-level embedding}\label{word_not_sentence}
We also studied the effect of using word-level embedding instead of sentence-level embedding in the retrieval of top k examples on both GPT 4 as well as on our model. Results of this can be found in Table \ref{word_level}.
\begin{table}[h]
  \scalebox{0.75}{
  \begin{tabular}{p{0.9\linewidth} | p{0.3\linewidth}}
    \toprule
    \textbf{Model}&\textbf{F1 score (\%)}\\
    \midrule
    IF-WRANER using sentence-level embedding & 75.72\\
    IF-WRANER using word-level embedding & 76.97\\
    RAG with GPT 4 using sentence-level embedding & 74.71\\
    RAG with GPT 4 using word-level embedding & 75.94\\
    \bottomrule
  \end{tabular}}
 \caption{\textmd{Effect of replacing sentence-level embedding with word-level embedding on model performance}}
 \label{word_level}
\end{table}

\subsection{Effect of changing the number of retrieved examples in LLM prompt}
We also studied the effect of changing the number of retrieved examples included in the prompt to IF-WRANER. Results of this can be found in Table \ref{ret_count}. We did not create separate deployments for testing the effect of varying number of retrieved examples. The latency numbers indicate the average time taken to complete one request to IF-WRANER on V100 GPU. Based on the trade-off between F1 score and latency we settled on 5 as the optimal number of retrieved examples to add to our LLM prompt.

\begin{table}[h]
  \scalebox{0.75}{
  \begin{tabular}{p{0.22\linewidth} | p{0.2\linewidth} | p{0.22\linewidth} | p{0.22\linewidth} | p{0.2\linewidth}}
    \toprule
    \textbf{Number of retrieved examples}&\textbf{F1 score on CrossNER(\%)}&\textbf{F1 score on Domain A(\%)} &\textbf{F1 score on Domain B(\%)} &\textbf{Median Latency (s)}\\
    \midrule
    1	& 65.32	&68.14	&64.98	&1.6\\
    3	& 74.28	& 76.25	& 72.49	& 1.8\\
    \textbf{5}	& \textbf{76.97}	& \textbf{83.72}	& \textbf{79.95} & \textbf{1.8}\\
    10	& 77.01	& 83.68	& 79.42	& 2.2\\
    20	& 76.88	& 84.29	& 79.27	& 2.8\\
    30	& 76.92	& 82.16	& 78.49	& 3.2\\
    \bottomrule
  \end{tabular}}
 \caption{\textmd{Effect of changing the number of retrieved examples included in LLM prompt}}
 \label{ret_count}
\end{table}

\subsection{Selecting the optimal vector DB and indexing scheme}
As mentioned in the paper, for the retrieval component, we have used Milvus DB as the vector DB for storing the word-level embeddings, IVF Flat as the indexing scheme and cosine similarity search for retrieving similar examples. We compared its performance against other vector DBs/vector similarity search libraries such as FAISS (\citep{douze2024faiss}) with different indexing schemes. A more detailed view of this can be found in Table \ref{vdb}. 

With flat indexing a direct comparison is made between embeddings whereas with IVF Flat indexing, embeddings of examples are clustered first and then the query embedding is compared with cluster embeddings. It is therefore faster. HSNW performs additional optimizations and is therefore even faster than IVF, but also shows a drop in F1 score. Performance of FAISS and Milvus are roughly similar, with Milvus showing slightly better numbers. Milvus with IVF Flat indexing provides a good tradeoff for our usecase and was therefore adopted.
\begin{table}[h]
  \scalebox{0.75}{
  \begin{tabular}{p{0.12\linewidth} |p{0.17\linewidth} | p{0.15\linewidth} | p{0.15\linewidth} | p{0.2\linewidth} | p{0.2\linewidth}}
    \toprule
    \textbf{Vector DB}&\textbf{Index}&\textbf{Similarity Function} &\textbf{F1 score on CrossNER(\%)} &\textbf{latency on 200 examples (s)} &\textbf{latency on $10^5$ examples (s)}\\
    \midrule
    Milvus	& Flat	   & cosine	  & 77.01	& 0.03	& 0.3\\
    \textbf{Milvus}	& \textbf{IVF Flat} & \textbf{cosine}	  & \textbf{76.97}	& \textbf{0.024}	& \textbf{0.1}\\
    FAISS	& FlatL2   & L2	      & 76.99	& 0.04	& 0.3\\
    FAISS	& IVF Flat & L2	      & 76.95	& 0.026	& 0.1\\
    FAISS	& HSNW	   & L2	      & 76.22	& 0.015	& 0.09\\
    \bottomrule
  \end{tabular}}
 \caption{\textmd{Comparison of different vector DBs and indexing schemes for use in the retrieval component of IF-WRANER. Search latency corresponds to median search latency.}}
 \label{vdb}
\end{table}

\subsection{Selecting the optimal embedder model}
While deciding on the retriever component, we also considered different embedder models. The table below compares different open-source embedder models, each using less than 2GB memory, for our use case. Models like text-ada-embedding, which only provide sentence-level embedding and abstract away the token level embedding vectors are excluded due to our focus being only on word-level embeddings.
Based on the experiments bge-base-en seems to work well across all domains.
\begin{table}[h]
  \scalebox{0.75}{
  \begin{tabular}{p{0.24\linewidth} | p{0.22\linewidth} | p{0.2\linewidth} | p{0.2\linewidth} | p{0.18\linewidth}}
    \toprule
    \textbf{Embedder Model}&\textbf{Memory requirement (GB)}&\textbf{CrossNER F1 score(\%)} &\textbf{Domain A F1 score(\%)} &\textbf{Domain B F1 score(\%)}\\
    \midrule
    \textbf{bge-base-en}	    & \textbf{1.63}	& \textbf{76.97}	& \textbf{83.72}	& \textbf{79.95}\\
    gte-large-en	& 1.62	& 76.88	& 83.75	& 78.86\\
    uae-large-v1	& 1.25	& 76.42	& 83.66	& 77.12\\
    \bottomrule
  \end{tabular}}
 \caption{\textmd{Comparison of different embedder models for the retrieval component of IF-WRANER}}
 \label{embedder}
\end{table}

\subsection{Selecting the optimal open-source LLM}\label{open_source_llm}
We experimented with different open-source LLMs such as 7B Meta LLM, 13B Meta LLM, Mistral-7B and so on. Based on performance and latency scores, we decided to use 7B Meta LLM for our experiments. We did not experiment with LLMs larger than 13B owing to latency and infra constraints. Details of this can be found in Table \ref{llms}.
\begin{table}[h]
  \scalebox{0.75}{
  \begin{tabular}{p{0.6\linewidth} | p{0.20\linewidth}| p{0.33\linewidth}}
    \toprule
    \textbf{Backbone LLM}&\textbf{F1 score (\%)}& \textbf{Latency on single V100 GPU (s)}\\
    \midrule
    \textbf{7B Meta LLM} (\citep{touvron2023llama}) & \textbf{76.97} & \textbf{1.8}\\
    13B Meta LLM (\citep{touvron2023llama}) & 77.42 & 2.6\\
    Mistral-7B (\citep{jiang2023mistral}) & 77.14 & 1.9\\
    Bloom-7b1 (\citep{le2023bloom}) & 74.42 & 1.9\\
    MPT-7B (\citep{MPT7b}) & 76.12 & 1.8\\
    \bottomrule
  \end{tabular}}
 \caption{\textmd{Comparison of different open-source LLMs}}
 \label{llms}
\end{table}

\subsection{Prompt Guidelines}
Based on our experimentation with different kinds of prompts and analysis of the model responses, we found that IF-WRANER demonstrates its best performance when the following prompting guidelines are followed:
\begin{itemize}[noitemsep,topsep=1pt,labelindent=0.5em,leftmargin=*]
    \item Entity definitions should be clear. Having atleast one example in the definition itself helps.
    \item When there is ambiguity between two entity types, such as ``Organization'' and ``Political Party'' and one is a subgroup of the other, then the subgroup entity type, in this case ``Political Party'', should appear first in the prompt. The model displays this behavior despite the regularization techniques applied to the model.
\end{itemize}

\end{document}